\documentclass[10pt,twocolumn,letterpaper]{article}

\usepackage[pagenumbers]{cvpr} 

\usepackage{graphicx}
\usepackage{amsmath}
\usepackage{framed}
\usepackage{amssymb}
\usepackage{booktabs}
\usepackage[normalem]{ulem}
\useunder{\uline}{\ul}{}

\usepackage[pagebackref,breaklinks,colorlinks]{hyperref}

\usepackage[capitalize]{cleveref}
\crefname{section}{Sec.}{Secs.}
\Crefname{section}{Section}{Sections}
\Crefname{table}{Table}{Tables}
\crefname{table}{Tab.}{Tabs.}

\def\cvprPaperID{*****} 
\def\confName{CVPR}
\def\confYear{2022}

\begin{document}

\title{Optimising 2D Pose Representation: Improve Accuracy, Stability and Generalisability Within Unsupervised 2D-3D Human Pose Estimation}

\author{Peter Hardy, Srinandan Dasmahapatra and Hansung Kim \\
University of Southampton\\
{\tt\small p.t.d.hardy@soton.ac.uk}
}
\maketitle
\begin{abstract}
This paper addresses the problem of 2D pose representation during unsupervised 2D to 3D pose lifting to improve the accuracy, stability and generalisability of 3D human pose estimation (HPE) models. All unsupervised 2D-3D HPE approaches provide the entire 2D kinematic skeleton to a model during training. We argue that this is sub-optimal and disruptive as long-range correlations are induced between independent 2D key points and predicted 3D ordinates during training. To this end, we conduct the following study. With a maximum architecture capacity of 6 residual blocks, we evaluate the performance of 5 models which each represent a 2D pose differently during the adversarial unsupervised 2D-3D HPE process. Additionally, we show the correlations between 2D key points which are learned during the training process, highlighting the unintuitive correlations induced when an entire 2D pose is provided to a lifting model. Our results show that the most optimal representation of a 2D pose is that of two independent segments, the torso and legs, with no shared features between each lifting network. This approach decreased the average error by 20\% on the Human3.6M dataset when compared to a model with a near identical parameter count trained on the entire 2D kinematic skeleton. Furthermore, due to the complex nature of adversarial learning, we show how this representation can also improve convergence during training allowing for an optimum result to be obtained more often. \textbf{Code and weights of our models will be made available}
\end{abstract}
\section{Introduction}
\label{sec:intro}
Monocular 3D human pose estimation (HPE) aims to reconstruct the 3D skeleton of the human body from 2D images or video. This is known to be an ill-posed inverse problem as multiple different 2D poses can correspond to the same 3D pose. Even with this hurdle, deep learning has allowed for accurate 2D-3D pose regression mappings to be learned allowing for remarkable results when trained and tested on 3D pose datasets \cite{Wandt2019RepNet, Martinez_2017_ICCV, Pavlakos2018CVPR, Yang20183DHP, Cheng_Yang_Wang_Tan_2020, Pavllo2019CVPR_Facebook}. Unfortunately, the difficulty of obtaining 3D datasets leads to poor performance when evaluating in domains where rigid environmental constraints (lighting, action, camera location, etc) are unable to be controlled. Recent work \cite{amazon_paper_2, Wandt_2022_CVPR, Drover_2018_ECCV_Amazon, Yu_2021_ICCV} has investigated if an unsupervised solution for 3D HPE is possible. These approaches utilize a geometric self-supervision cycle through random rotations to create a consistent lifting network, with some form of pose or probability discriminator to see if the rotated pose, once reprojected back to 2D, is realistic. As 2D data is cheaper to obtain, more efficient for computations and readily available in many circumstances, improving the performance of unsupervised 2D-3D HPE networks would therefore allow for accurate 3D poses to be obtained in many unconstrained scenarios.

An overlooked aspect in prior work however, is the representation of the 2D pose being given to the lifting model. We posit that when a full 2D pose is provided to a lifting model during training, long-range correlations are induced between a key points 3D prediction and all of the poses' other 2D key point coordinates (i.e. the 3D prediction of the left elbow will be influenced by the 2D coordinate of the right ankle). Although supervised approaches have touched upon this topic by learning joint dependency through graph convolutional networks (GCN) \cite{Kyoungoh_2018_ECCV, Zhao_2019_CVPR}, relationships between joints via relational networks \cite{Park_2018_BMVC} or splitting and recombining the limbs of a pose \cite{Ailing_ECCV_2020_SRNET}. To our best knowledge, this has never been done in an unsupervised setting, and it has never been assumed that the pose could be two or more independent structures with no inter-relational correspondence needed. Additionally, the large variations in network architecture and optimization within prior work mean we are unable to fairly compare the results between approaches to find an optimum representation.

We address this problem by training 5 models with near identical amounts of parameters and identical training approaches on different 2D pose representations, to accurately obtain a 3D pose. By evaluating the results obtained from each model we will be able to determine an optimum representation that future work can use to obtain the best performance. We will also show the correlations induced between 2D key points during training when a full pose is provided to a model as well as our best performing 2D representation model, providing some intuition behind the improved performance. Furthermore, we will touch upon improving this model further by enabling it to learn what it thinks to be the best 2D representation of a pose. To summarise our paper makes the following contributions:
\begin{itemize}
    \item We show the effect that different 2D pose representations have on the unsupervised adversarial 2D-3D HPE process, where utilizing the best 2D pose representation can reduce the average error by 20\%.
    \item Our findings can be easily implemented within current the state of the art as our approach utilizes the popular residual block introduced by Martinez \etal \cite{Martinez_2017_ICCV} and used within \cite{amazon_paper_2, Wandt_2022_CVPR, Wandt2019RepNet, Drover_2018_ECCV_Amazon, Yu_2021_ICCV}.
    \item We show the correlations induced between key points for a full 2D pose representation model and our best 2D pose representation model highlighting the sub-optimal learning when a full 2D pose is provided to a network. 
    \item We show the adversarial stability of our best pose representation model against a full 2D pose representation model, highlighting that the improvement is consistent with multiple random initializations.
\end{itemize}


\section{Related Work}
There currently exist two main avenues of deep-learning research for 3D HPE. The first learns the mapping of 3D joints directly from a 2D image \cite{Pavlakos2017CoarsetoFineVP, 8795015, mono-3dhp2017, Li2015MaximumMarginSL, TomeD2017LftD}. The second builds upon an accurate intermediate 2D pose estimate, with the 2D pose obtained from an image through techniques such as Stacked-Hourglass Architectures \cite{stacked_hourglass} or Part Affinity Fields \cite{open_pose}, and lifts this 2D pose to 3D. This work focuses on the latter 2D to 3D lifting avenue which can be organized into the following categories:

\subsection{Fully Supervised}

Fully supervised approaches seek to learn mappings from paired 2D-3D data which contain ground truth 2D locations of key points and their corresponding 3D coordinates. Martinez et al. \cite{Martinez_2017_ICCV} introduced a baseline fully connected regression model which learned 3D coordinates from their relative 2D locations. Exemplar approaches such as Chen et al. \cite{Chen20173DHP} and Yang et al. \cite{YANG2019371} use large dictionaries/databases of 3D poses with a nearest-neighbor search to determine an optimal 3D pose. Pavllo et al. \cite{Pavllo2019CVPR_Facebook} used temporal convolutions over 2D key points to predict the pose of the central or end frame in a time series, whereas Mehta et al. \cite{VNect_SIGGRAPH2017} utilized multi-task learning to combine a convolutional pose regressor with kinematic skeleton fitting for real-time 3D HPE.

\subsection{Weakly-Supervised}

Weakly-Supervised approaches do not use explicit 2D-3D correspondences and instead use either augmented 3D data during training or unpaired 2D-3D data to learn human body priors (shape or articulation). Pavlakos et al. \cite{Pavlakos} and Ronchi et al. \cite{ordinal_paper_2} proposed learning of 3D poses from 2D with ordinal depth relationships between key points (e.g. the right wrist is behind the right elbow). Wandt and Rosenhahn \cite{Wandt2019RepNet} introduced a weakly-supervised adversarial approach where they transformed their predicted and ground truth 3D poses into a kinematic chain \cite{Wandt2018AKC} before being seen by a Wasserstein critic network \cite{improved_wasserstein}. Yang et al. \cite{Yang20183DHP} lifted 2D poses from in the wild scenes where no ground truth data was available with a critic network that compared these predictions against ground-truth 3D skeletons. By contrast, Zhou et al. \cite{Zhou_2017_ICCV} did not use a critic network but transfer learning that used mixed 2D and 3D labels in a unified network to predict 3D poses in scenarios where no 3D data is available. Drover et al. \cite{Drover_2018_ECCV_Amazon} investigated if 3D poses can be learned through 2D self-consistency alone, where they would rotate a predicted 3D pose and reproject it back into 2D before passing it back to the model for comparison. This led to the discovery that a 2D critic network was also needed and self-consistency alone is not sufficient.

\subsection{Unsupervised}

Unsupervised approaches do not utilize any 3D data during training, unpaired or otherwise. Kudo et al. \cite{kudo2018unsupervised} introduced one of the first unsupervised adversarial networks which utilized random reprojections and a 2D critic network, under the assumption that any predicted 3D pose once rotated and reprojected should still produce a believable 2D pose. Chen et al. \cite{amazon_paper_2} expanded this work and that of \cite{Drover_2018_ECCV_Amazon} by introducing an unsupervised adversarial approach with a self-consistency cycle. They also provided ablation studies highlighting a 7\% improvement when using temporal cues during training. Yu et al. \cite{Yu_2021_ICCV} built upon \cite{amazon_paper_2} highlighting that these temporal constraints may hinder a models performance due to balancing multiple training objectives simultaneously and proposed splitting the problem into both a lifting and scale estimation module. They also found that adding temporal motion consistency can boost the performance of their model by 6\%.

\subsection{Prior Work on 2D Pose Representations}

Park et al. \cite{Park_2018_BMVC} introduced one of the first studies that investigated if the human pose could be constructed of multiple interrelated structures. By utilizing relational networks, which were passed limb pairs during training ([left arm and right arm], [left arm and right leg], etc.), they aimed to discover shared feature information between limbs that would aid in the 2D-3D lifting process. Kyoungoh et al. \cite{kyoungoh} utilized a propagating LSTM architecture where each LSTM would sequentially add another key point to its input. This intrinsic approach allowed joint relationships to be learnt0 over time as each key point was added. Jiang Hao \cite{exemplar_upper_lower} worked on an exemplar approach where their 3D dictionary was split into torso and legs key points during lookup. Though their idea behind this was to speed up the nearest-neighbor search process, they may have improved the generalisability of their model as it could find similar legs in one training example and similar torsos within another, increasing the number of possible library combinations. Zeng et al. \cite{Ailing_ECCV_2020_SRNET} investigated splitting a pose into different parts to learn localized correlations between key points first then global correlations between these localized segments afterward. Unlike our study however, \cite{Ailing_ECCV_2020_SRNET} is a supervised network that still assumes that an entire 2D pose is one independent structure via feature sharing between localized groups throughout their network. Our study utilizes completely unsupervised networks, where two of the pose representations we investigate utilize no feature sharing, with the idea that localized key-point groups can be assumed to be independent of one another.


\section{Methodology}

In this section, we first describe the different 2D representations used and the intuition behind each representation. This is followed by an explanation of the adversarial learning approach of our various models.

\subsection{2D Pose Representations}

Our 2D poses consist of $N$ key points $(x_i, y_i)$, $i = 1 ...N$, with the root key point, the midpoint between the left and right hip joint, being located at the origin $(0, 0)$. Because of the fundamental scale ambiguity associated with monocular images, it is impossible to infer absolute depth from a single view alone \cite{depth}. Therefore, we used max-normalization on each of our 2D poses to scale their 2D key point coordinates between -1 and 1. In our study we investigate using the following five 2D pose representations and how they affect the unsupervised 2D-3D HPE process:
\begin{itemize}
    \item \textbf{Full 2D Pose Representation:} The standard approach where a model is taught to predict the 3D ordinates from the entire 2D pose. The theory being that a good model should learn to correlate co-dependent key points and disassociate independent ones. In this scenario, we will have one network take an entire 2D pose as input and predict the 3D ordinates for the full pose.
    
    \item \textbf{Split and Recombine Leg and Torso key points:} A concept introduced by \cite{Ailing_ECCV_2020_SRNET}, the idea behind this representation is that we should first learn localized key point correlations and then the global context of a pose. In this scenario, two separate networks will first learn the localized correlations between the leg and torso key points and output a vector of features learned. A further network will learn to combine these features and predict the 3D ordinates for the full pose.
    
    \item \textbf{Independent Leg and Torso key points:} Although there is intuition behind first learning local and then global context. For many actions (e.g. waving, eating, etc) the legs and torso of a person may be more independent than co-dependent. To see if this is true another 2D pose representation will be that of completely independent torso and leg key points. In this scenario, we will have two independent lifting networks one that accepts the 2D key points and predicts the 3D ordinates for the torso and the other the 2D key points and 3D ordinates of the legs.
    
    \item \textbf{Split and Recombine 5 Limbs key points:} Although many human actions require both hands or legs to move in sync, in reality, humans can easily move each limb independently of one another. In this approach, 5 separate networks will learn the localized correlations of each limb (left arm, right arm, torso, right leg and left leg) key points and output a vector of features learned. A further network will learn to combine these features and predict the 3D ordinates for the full pose.
    
    \item \textbf{Independent 5 Limbs key points:} Similar to the Independent leg and torso representation, this approach will look at the possibility of treating the entire pose as 5 independent limbs. In this scenario, 5 separate networks will each take as input and predict the 3D ordinates for each limb independently (i.e. one network will take 3 key points as input belonging to the right arm and predict the 3D ordinates for the right arm).
\end{itemize}

\subsection{Lifting Network}
The lifting networks ($G$) used are fully connected neural networks whose architecture was based on \cite{Martinez_2017_ICCV} and can be seen in Figure \ref{fig:residualblock} in Appendix \ref{architecture_figures}. These predict one 3D ordinate relative to the root for each 2D key point provided:
\begin{align}
\label{generator_function}
G(\mathbf{x}, \mathbf{y}; \mathbf{w}) = \hat{\mathbf{z}}
\end{align}
where $(\mathbf{x}, \mathbf{y})$  are the 2D coordinates provided to the network, $\hat{\mathbf{z}}$ are the predicted 3D ordinate for each key point and $\mathbf{w}$ are the weights of our model learned during training. Once our lifting networks had made their predictions they are concatenated with the original 2D key points to create our final predicted 3D pose $(\mathbf{x}, \mathbf{y}, \hat{\mathbf{z}})$. A more detailed architecture explanation of each pose representation model can be seen in Appendix \ref{architecture_figures}.

\subsection{Reprojection Consistency} \label{reprojection_loss_explained}
Similar to prior work \cite{amazon_paper_2, Drover_2018_ECCV_Amazon, Yu_2021_ICCV}, we utilize a self-consistency cycle through random 3D rotations to reproject our predicted 3D poses to new synthetic 2D viewpoints. Let $\mathbf{Y}\in\mathbb{R}^{N \times 2}$ be our full 2D pose. Once a prediction $G(\mathbf{Y})$ is made and a full 3D pose $(\mathbf{x}, \mathbf{y}, \hat{\mathbf{z}})$ obtained, a random rotation matrix $\mathbf{R}$ will be created by uniformly sampling an azimuth angle between [$-\frac{8\pi}{9}, \frac{8\pi}{9}$] and an elevation angle between [$\frac{-\pi}{18}, \frac{\pi}{18}$]. The predicted 3D pose will be rotated by this matrix and reprojected back to 2D via projection $\mathbf{P}$, obtaining a new synthetic viewpoint of the pose and the matrix $\tilde{\mathbf{Y}}\in\mathbb{R}^{N \times 2}$ where $\tilde{\mathbf{Y}} = \mathbf{PR}[(\mathbf{x}, \mathbf{y}, \hat{\mathbf{z}})]$. To enable our model to learn consistency, if we now pass $\tilde{\mathbf{Y}}$ to the same lifting network, perform the inverse rotation $\mathbf{R}^{-1}$ on the newly predicted 3D pose $(\tilde{\mathbf{x}}, \tilde{\mathbf{y}}, \tilde{\hat{\mathbf{z}}})$ and reproject it back into 2D, we should obtain our original matrix of 2D key points $\mathbf{Y}$. This cycle allows our lifting networks to learn self-consistency during training where they seek to minimize the following components in the loss function:
\begin{align}
\mathcal{L}_{2D} = \|\mathbf{Y} - \mathbf{PR}^{-1}[G(\tilde{\mathbf{Y}})]\|^2
\end{align}
In the case of our independent leg and torso and independent 5 limb 2D pose approaches, each network will receive its own $\mathcal{L}_{2D}$ loss based on the error between the key points that they predicted. As an example, part of the $\mathcal{L}_{2D}$ loss for the right arm lifter within our independent 5 limbs representation would include the difference between the original 2D key point coordinate of the right wrist, and its 2D coordinate once $\tilde{\textbf{Y}}$ was inversely rotated and reprojected. This error would not be included in the $\mathcal{L}_{2D}$ loss for the left arm lifter as it does not predict the 3D ordinate for this key point.

\subsection{90 Degree Consistency}
\begin{figure}[!htb]
\begin{centering}
\includegraphics[width=\linewidth]{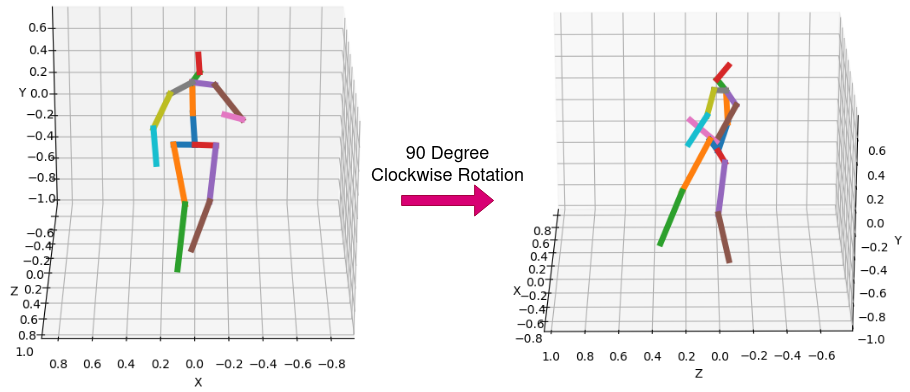}
\caption{Showing that a $90^\circ$ rotation of a 3D pose around the $y$ axis with a fixed camera position, will result in the $x$ axis values of the pose prior to rotation representing the $z$ axis values of the pose after the rotation and vice versa.}
\label{fig:90degree}
\end{centering}
\end{figure}

During our study, we found that increasing self-consistency was key to reducing the evaluation error (see Appendix \ref{consistency_cycle}). Therefore, we introduce new self-consistency constraints during training based on rotations around the $y$ axis at $90^\circ$ increments. Let $(\mathbf{x}, \mathbf{y}, \hat{\mathbf{z}})$ be the predicted 3D pose from our model. If we assume a fixed camera position and rotate our pose $90^\circ$, then the depth component of our pose ($\mathbf{\hat{z}})$ prior to rotation will now lie on the $x$ axis from our cameras viewpoint. A visual example of this can be seen in Figure \ref{fig:90degree}.
As we have normalized the axis of our poses between -1 and 1, a $90^\circ$ clockwise rotation of the 3D pose $(\mathbf{x}, \mathbf{y}, \hat{\mathbf{z}})$ will produce the pose $(\hat{\mathbf{z}}, \mathbf{y}, -\mathbf{x})$. Therefore, providing $(\hat{\mathbf{z}}, \mathbf{y})$ as input to our generators should result in $-\mathbf{x}$ as its predictions. This fact allows for the inclusion of three additional consistency constraints in the loss function of our generators which teach consistency at a $90^\circ$ clockwise rotation, a $90^\circ$ anticlockwise rotation and a $180^\circ$ rotation, which are as follows:
\begin{align}
\|G({\hat{\mathbf{z}}}, \mathbf{y}; \mathbf{w}) + \mathbf{x}\|^2 = 0
\end{align}
\begin{align}
\|G({-\hat{\mathbf{z}}}, \mathbf{y}; \mathbf{w}) - \mathbf{x}\|^2 = 0
\end{align}
\begin{align}
\|G(\mathbf{x}, \mathbf{y}; \mathbf{w}) + G(-\mathbf{x}, \mathbf{y}; \mathbf{w})\|^2 = 0
\end{align}
The left parts of these constraints are summed in the final loss function to produce $\mathcal{L}_{90^\circ}$ and included in the optimization function of our models. Similar to $L_{2D}$ in the case of our independent 2D pose representation networks, each lifting network will have its own $L_{90^\circ}$ depending on the 2D key points that it predicted for. Although we could include three similar constraints for $90^\circ$ rotational increments around the $x$ axis, we found that these hinder the performance of the model. This is due to $90^\circ$ $x$ axis rotations producing a birds eye and ground-up view of a 2D pose, which contains little variation between 2D key points.

\subsection{Discriminator Loss}
Although self-consistency is important, alone it is insufficient for generating realistic 3D skeletons \cite{amazon_paper_2}. Therefore, we utilize a 2D discriminator $D$, that takes as input a 2D pose and outputs a probability value of that pose being plausible or unrealistic. The architecture of our discriminator is a fully connected neural network made of 3 residual blocks \cite{Martinez_2017_ICCV} and ending with a linear prediction function. It learns to discriminate between the real 2D poses within our data $\mathbf{Y}$, and our reprojected 2D pose $\tilde{\mathbf{Y}}$. This provides feedback to the generators during training, enabling the learning of geometric priors such as joint angles and limb length ratios. Our discriminator $D$ utilized the least squares loss \cite{LSGAN} as we found it performed better than the standard adversarial loss during our experiments, and defined as:
\begin{equation}
\begin{aligned}
  \mathop{\mbox{min}}_{D} &= \frac{1}{2}\mathbb{E}[D(\mathbf{Y})-1]^2 + \frac{1}{2}\mathbb{E}[D(\tilde{\mathbf{Y}})]^2  \\
  \mathop{\mbox{min}}_{G} &= \frac{1}{2}\mathbb{E}[D(\tilde{\mathbf{Y}}) -1]^2 
\end{aligned}
\end{equation}
Unlike the consistency constraints, we do not provide a unique version of $L_{adv}$ to each lifting network in our independent 2D pose representation networks and instead provide the same loss (with a different weight) to each network. This is due to two reasons; Firstly, although our independent lifting networks are being trained separately, we still want them to produce a believable reprojected 2D pose together and having one discriminator see the entire pose will provide this feedback during training. Secondly, we found that trying to discriminate between segments of a 2D pose provided poor feedback during training. An example being that 2D legs and arms are normally represented as straight or bent lines, making it hard for a discriminator to tell plausible and implausible limbs apart.

\subsection{Training}
As discussed, our lifting networks were trained adversarially with random reprojection and $90^\circ$ consistency constraints. The network parameters are then updated to optimize the total loss for each lifting model given by:
\begin{equation}
    \mathcal{L} = w_1\mathcal{L}_{adv} + w_2\mathcal{L}_{2D} + w_3\mathcal{L}_{90^\circ}
\end{equation}
where, $w_1$, $w_2$ and $w_3$ are the relative weights for the adversarial loss component, reprojection consistency loss component and $90^\circ$ consistency loss component respectively. The numerical values of the weights for each network are defined in Appendix \ref{architecture_figures}. We trained our models completely unsupervised following \cite{amazon_paper_2} using a batch size of 8192 and the Adam optimizer \cite{Kingma2015AdamAM} with a learning rate of 0.0002. Our experiments use $N=16$ key points. During discriminator training we employ label flipping with a 10\% chance.
\section{Evaluation and Results}
Here we compare and evaluate the performance of our different 2D pose representation models against each other and various state-of-the-art models. Though our aim is not to be or come close to the current state of the art, we believe that comparing against current models is important to examine how just a simple change in 2D pose representation can compare against more complex architectures and approaches. Our results show that not only can an optimum 2D pose representation improve upon the default full pose representation by 20\%, but it can also improve generalisability to unseen poses, even when compared against more complex supervised and weakly-supervised approaches.

\subsection{Quantitative Results On Human3.6M}
Human3.6M (H36M) \cite{h36m_pami} is one of the largest and most widely used 3D human pose datasets, containing 3.6 million 3D human poses. It consists of both video and motion capture (MoCap) data from 4 viewpoints of 5 female and 6 male subjects performing specific actions (e.g. talking on the phone, taking a photo, eating, etc.). There are two main evaluation protocols for the H36M dataset, which use subjects 1, 5, 6, 7 and 8 for training and subject 9 and 11 for evaluation. Both protocols report the Mean Per Joint Position Error (MPJPE), which is the Euclidean distance in millimeters between the predicted and ground truth 3D coordinates. We report the protocol-II performance of our model on all frames of the validation set which employs rigid alignment between the ground truth and predicted pose before evaluation. The results of our different 2D pose representation models can be seen in Table \ref{H36M_Results_Table}.
\begin{table*}[!htb]
\centering
\resizebox{0.92\textwidth}{!}{%
\begin{tabular}{@{}llcccccccc@{}}
\toprule
Method                                         & Approach          & Direct.       & Discuss       & Eat           & Greet         & Phone         & Photo         & Posing        & Purchase      \\ \midrule
Martinez et al. \cite{Martinez_2017_ICCV} (GT)                           & Supervised        & 39.5          & 43.2          & 46.4          & 47.0          & 51.0          & 56.0          & 41.4          & 40.6          \\
Pavllo et al. \cite{Pavllo2019CVPR_Facebook} (GT)                             & Supervised        & 36.0          & {\ul 38.7}    & 38.0          & 41.7          & {\ul 40.1}    & {\ul 45.9}    & 37.1          & 35.4          \\
Cai et al. \cite{cai} (GT)                                & Supervised        & 36.8          & {\ul 38.7}    & 38.2          & 41.7          & 40.7          & 46.8          & 37.9          & 35.6          \\
Yang et al. \cite{Yang20183DHP} (+)                                & Weakly-Supervised & \textbf{26.9} & \textbf{30.9} & {\ul 36.3}    & 39.9          & 43.9          & 47.4          & \textbf{28.8} & \textbf{29.4} \\
Wandt and Rosenhahn \cite{Wandt2019RepNet} (GT)                       & Weakly-Supervised & {\ul 33.6}    & 38.8          & \textbf{32.6} & \textbf{37.5} & \textbf{36.0} & \textbf{44.1} & 37.8          & {\ul 34.9}    \\
Pavlakos et al. \cite{Pavlakos2018CVPR} (+)                            & Weakly-Supervised & 34.7          & 39.8          & 41.8          & {\ul 38.6}    & 42.5          & 47.5          & 38.0          & 36.6          \\
Chen et al. \cite{amazon_paper_2} (GT)(T)                            & Unsupervised      & -             & -             & -             & -             & -             & -             & -             & -             \\
Yu et al. \cite{Yu_2021_ICCV} (GT)(T)                              & Unsupervised      & -             & -             & -             & -             & -             & -             & -             & -             \\ \midrule
Full 2D Pose Network (Ours)(GT)                & Unsupervised      & 51.5          & 47.1          & 48.1          & 53.7          & 47.6          & 56.2          & 45.9          & 46.1          \\
Split-Combine Leg and Torso Network (Ours)(GT) & Unsupervised      & 45.3          & 44.8          & 43.4          & 47.2          & 46.2          & 48.2          & 43.2          & 41.5          \\
Independent Leg and Torso Network (Ours)(GT)   & Unsupervised      & 38.1          & {\ul 38.7}    & 36.8          & 42.0          & 41.9          & 49.9          & {\ul 36.8}    & 38.8          \\
Split-Combine 5 Limbs Network (Ours)(GT)       & Unsupervised      & 43.6          & 42.3          & 43.4          & 45.9          & 45.1          & 49.9          & 41.5          & 42.0          \\
Independent 5 Limbs Network (Ours)(GT)         & Unsupervised      & 90.0          & 98.1          & 109.4         & 94.5          & 100.3         & 109.3         & 74.0          & 105.5   \\ \bottomrule
\end{tabular}%
}
\centering
\resizebox{0.92\textwidth}{!}{%
\begin{tabular}{@{}llccccccc|c@{}}
\toprule
Method                                         & Approach          & Sit           & SitD.         & Smoke         & Wait          & Walk          & WalkD.        & WalkT.        & Avg.          \\ \midrule
Martinez et al. \cite{Martinez_2017_ICCV} (GT)                           & Supervised        & 56.5          & 69.4          & 49.2          & 45.0          & 49.5          & 38.0          & 43.1          & 47.7          \\
Pavllo et al. \cite{Pavllo2019CVPR_Facebook} (GT)                             & Supervised        & 46.8          & 53.4          & 41.4          & 36.9          & 43.1          & \textbf{30.3} & 34.8          & 40.0          \\
Cai et al. \cite{cai} (GT)                                & Supervised        & 47.6          & \textbf{51.7} & 41.3          & {\ul 36.8}    & 42.7          & {\ul 31.0}    & {\ul 34.7}    & 40.2          \\
Yang et al. \cite{Yang20183DHP} (+)                                & Weakly-Supervised & \textbf{36.9} & 58.4          & 41.5          & \textbf{30.5} & \textbf{29.5} & 42.5          & \textbf{32.2} & \textbf{37.7} \\
Wandt and Rosenhahn \cite{Wandt2019RepNet} (GT)                       & Weakly-Supervised & {\ul 39.2}    & {\ul 52.0}    & \textbf{37.5} & 39.8          & 34.1          & 40.3          & 34.9          & {\ul 38.2}    \\
Pavlakos et al. \cite{Pavlakos2018CVPR} (+)                            & Weakly-Supervised & 50.7          & 56.8          & 42.6          & 39.6          & 43.9          & 32.1          & 36.5          & 41.8          \\
Chen et al. \cite{amazon_paper_2} (GT)(T)                            & Unsupervised      & -             & -             & -             & -             & -             & -             & -             & 51.0          \\
Yu et al. \cite{Yu_2021_ICCV} (GT)(T)                              & Unsupervised      & -             & -             & -             & -             & -             & -             & -             & 42.0          \\ \midrule
Full 2D Pose Network (Ours)(GT)                & Unsupervised      & 55.4          & 63.9          & 47.0          & 47.5          & 46.5          & 50.7          & 46.6          & 50.0          \\
Split-Combine Leg and Torso Network (Ours)(GT) & Unsupervised      & 55.3          & 66.9          & 45.2          & 43.5          & 39.8          & 47.4          & 43.1          & 46.8          \\
Independent Leg and Torso Network (Ours)(GT)   & Unsupervised      & 47.4          & 60.6          & {\ul 40.6}    & 38.0          & {\ul 34.0}    & 46.0          & 37.1          & 41.7          \\
Split-Combine 5 Limbs Network (Ours)(GT)       & Unsupervised      & 50.1          & 63.4          & 43.3          & 42.6          & 42.6          & 46.9          & 42.4          & 45.5          \\
Independent 5 Limbs Network (Ours)(GT)         & Unsupervised      & 123.1         & 154.6         & 96.6          & 86.1          & 87.4          & 108.4         & 96.0          & 102.2         \\ \bottomrule
\end{tabular}%
}
\caption{The reconstruction error (MPJPE) on the H36M Validation Set. \textbf{Legend}: (+) denotes extra data during training. (GT) denotes providing 2D ground truth key points to a network for prediction. (T) denotes the use of temporal information. All compairson results are taking from their respective papers. Lower is better, best in bold, second best underlined.}
\label{H36M_Results_Table}
\end{table*}
From our results, we can see that representing a 2D pose as an independent torso and legs improves performance the most, as this representation achieved a 20\% lower average error than our full 2D pose network. Additionally, our split-combine leg and torso and split-combine 5 limbs representation also showed improved performance, decreasing the average error by 6\% and 9\% respectively. Our independent 5 limbs model did not improve results which may be due to this representations models using 3 or 4 key points to predict 3D ordinates which may not contain enough information to learn the task at hand. By looking at our leg and torso results we can see it additionally performed well against several fully supervised models. In fact, by using this representation we achieved the second highest performance in the Discussing, Posing, Smoking and Walking action. Although improving the Discussing, Posing and Smoking action seems intuitive, as we would assume the arms and legs of a person to be independent during these actions, it is surprising that we achieved such high results in walking as we would expect to see larger levels of co-dependence between the torso and legs during this movement.
\subsection{Quantitative Results On MPI-INF-3DHP} 
MPI-INF-3DHP \cite{mono-3dhp2017} is a markerless MoCap dataset containing the 3D human poses of 8 actors performing 8 different activities. We use this dataset to highlight the effect that 2D pose representations have on a model's generalisability to unseen poses when trained on the H36M dataset. The evaluation metrics used are the \emph{percentage of correctly positioned key points} (PCK3D) and \emph{area under the curve} (AUC) as defined by \cite{mono-3dhp2017}. As our predicted poses are normalized, we upscale their 3D coordinates by the original 2D normalizing factor before evaluation. Our results can be seen in Table \ref{mpi_inf_results}.
\begin{table}[]
\centering
\resizebox{0.99\linewidth}{!}{%
\begin{tabular}{@{}llcc@{}}
\toprule
Method                                           & Approach          & PCK3D & AUC  \\ \midrule
Mehta et al. \cite{mono-3dhp2017} (3DHP)(H36M)(*)                    & Supervised        & 76.5  & 40.8 \\
Zeng et al. \cite{Ailing_ECCV_2020_SRNET} (H36M)                               & Supervised        & 77.6  & 43.8 \\
Yang et al. \cite{Yang20183DHP} (H36M)(+)                            & Weakly-Supervised & 69.0  & 32.0 \\
Wandt and Rosenhahn \cite{Wandt2019RepNet} (H36M)                       & Weakly-Supervised & {\ul 81.8}  & \textbf{54.8} \\
Kanazawa et al. \cite{kanazawa} (3DHP)(T)                        & Weakly-Supervised & 77.1  & 40.7 \\
Chen et al. \cite{amazon_paper_2} (3DHP)(T)                            & Unsupervised      & 71.1  & 36.3 \\
Kundu et al. \cite{Kundu2020KinematicStructurePreservedRF} (H36M)                              & Unsupervised      & 76.5  & 39.8 \\
Yu et al. \cite{Yu_2021_ICCV} (H36M)(T)                              & Unsupervised      & \textbf{82.2}  & 46.6 \\ \midrule
Full 2D Pose Network (H36M)(Ours)                & Unsupervised      & 75.4  & 45.8 \\
Split-Combine Leg and Torso Network (H36M)(Ours) & Unsupervised      & 76.1  & 47.5 \\
Independent Leg and Torso Network (H36M)(Ours)   & Unsupervised      & 78.5  & {\ul 48.5} \\
Split-Combine 5 Limbs Network (H36M)(Ours)       & Unsupervised      & 77.7  & 48.2 \\
Independent 5 Limbs Network (H36M)(Ours)         & Unsupervised      & 47.2  & 25.6 \\ \bottomrule
\end{tabular}%
}
\caption{Results on the MPI-INF-3DHP dataset. Our models are only trained on H36M Legend: (3DHP) denotes a model being trained
on the MPI-INF-3DHP dataset. (H36M) denotes a model being trained on the Human3.6M dataset. (+) denotes additional training data.
(*) uses transfer learning during from 2Dposenet. (T) denotes the use of temporal information during training. All comparison results are
taking from their respective papers. Higher is better, best in bold, second best underlined.}
\label{mpi_inf_results}
\end{table}
Similar to our H36M results, all of our 2D representation approaches, aside from our independent 5 limbs representation, performed better on unseen poses when compared to a full 2D pose representation approach. Our independent leg and torso representation improved results the most with a 4\% increase in PCK3D and 6\% increase in AUC when compared to the full 2D pose network. Additionally, this representation achieved the second highest overall AUC metric, even compared to other supervised and weakly-supervised approaches. 
Our reasoning for improved generalisability stems from the individualistic nature of our independent leg and torso model. For example, let us say we have an unseen 2D pose we want to lift into 3D $\tilde{\textbf{Y}}$ which happens to contain identical 2D leg key points to pose $\textbf{Y}_A$ and identical 2D torso key points to pose $\textbf{Y}_B$ which were both in the training data. The leg network during training will have seen and learned to predict the 3D ordinates of the leg-keypoints within $\textbf{Y}_A$ without needing the torso key points. Similarly, the torso network saw the torso key points within $\textbf{Y}_B$ and learned to predict these 3D ordinates with no knowledge of the leg key points. Therefore this new example $\tilde{\textbf{Y}}$ is not a difficult pose to predict the 3D ordinates for as the model has seen both segments of the pose before just within different training examples. Our full 2D pose by contrast saw both the leg and torso key points of $\textbf{Y}_A$ and $\textbf{Y}_B$ during training and learned to correlate them together during the 2D-3D lifting process. Meaning that when we now provide it with $\tilde{\textbf{Y}}$ it struggles to predict accurate 3D ordinates as though it has seen parts of this pose before the entire pose as a whole is new. We believe this also explains why our independent model had higher generalisability performance than both our split-recombine approaches. Although features are first obtained from localized groups in these approaches, the combination of these localized groups' features leads to a poorer generalizing model as it has learned to correlate sub-structures together during prediction.
\section{Reducing unintuitive key point correlations}
\begin{figure}[!htb]
\centering
\includegraphics[width=0.85\linewidth]{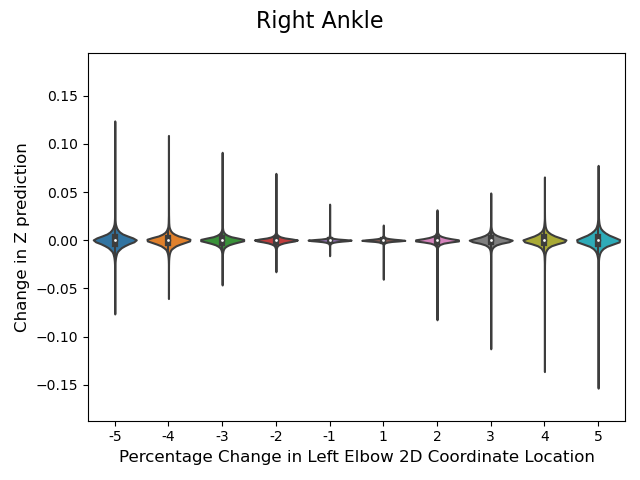}\hfill \\
\includegraphics[width=0.85\linewidth]{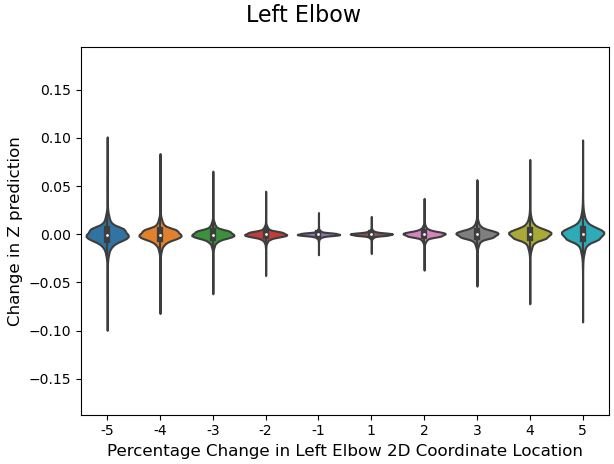}
\caption{Showing the effect that changing the left elbows 2D key point coordinates has on the 3D ordinate prediction of the right ankle (top) and the 3D ordinate prediction of the left elbow (bottom) for the full 2D pose representation model.}
\label{fig:correlation figure}
\end{figure}

\begin{figure}[!htb]
\centering
\includegraphics[width=0.85\linewidth]{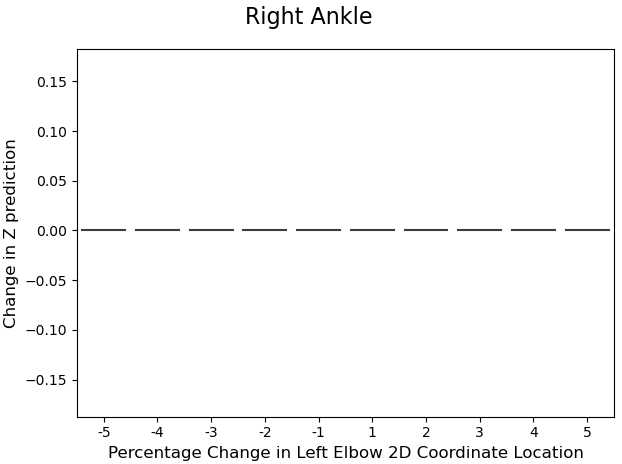}\hfill \\
\includegraphics[width=0.85\linewidth]{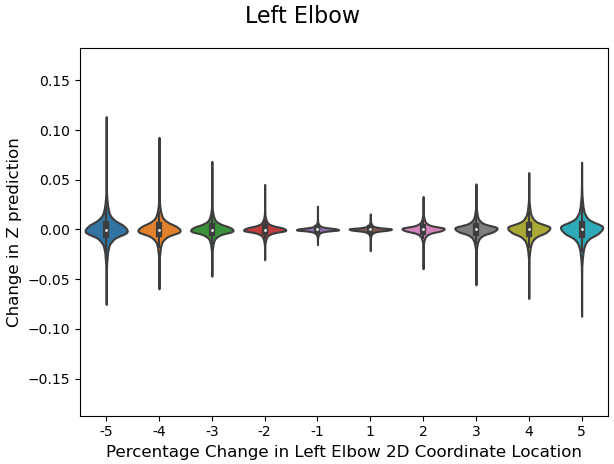}\hfill
\caption{Showing the effect that changing the left ankles 2D key point coordinates has on the 3D ordinate prediction of the left ankle for the full 2D pose model (top) and the independent leg and torso model (bottom).}
\label{fig:correlation figure 2}
\end{figure}

Our intuition behind why we obtain better results when using an independent leg and torso network when compared to a full 2D pose network, is due to the model learning more intuitive correlations between 2D key points and 3D ordinates. To validate our assumption we performed the following additional study. By changing the coordinate value of a 2D key point within a 2D pose, we observed how the 3D predictions of our models are affected and from this see if any strange correlations have been made. For example, if we changed the key point location for the left knee and the largest change in 3D prediction was for the right shoulder, then we can say that the model has learned to correlate the 2D key point of the left knee with its 3D prediction of the right shoulder. To conduct this study each 2D key point within the H36M validation set was scaled in 1\% increments between -95\% to 105\% in size. Each key point was changed separately and we observed the change in prediction for each 3D ordinate caused by changing that individual key point. Figures \ref{fig:correlation figure} and \ref{fig:correlation figure 2} provide a sample of the results from this study.

Examining Figure \ref{fig:correlation figure} we can see that the full 2D pose representation model has learned to correlate any change in the 2D coordinate for the left elbow with the 3D ordinate prediction of the right ankle and left elbow equally. In fact, where the left elbow coordinate increased by 5\% the 3D ordinate prediction of the deviation in the right ankle is larger than the left elbow's 3D ordinate prediction. This highlights the unintuitive correlations learned when a model is trained on the entire human pose, as humans are able to move their left elbow without moving their right ankle though the model has failed to learn this. Figure \ref{fig:correlation figure 2} contains our independent leg and torso model results where the correlation between the left elbow and right ankle is now removed, while maintaining the correlation between the left elbows 2D coordinate and 3D ordinate prediction. We believe that these more intuitive correlations is the reasoning behind our improved results as our independent models predictions are less sensitive to a change in a specific key point. The full results of this study can be seen in Figure \ref{fig:Left Ankle correlations} - \ref{fig:crown correlations 2} in the appendix.
\section{Improving Adversarial Training Stability}
\begin{figure}
\centering
\includegraphics[width=0.95\linewidth]{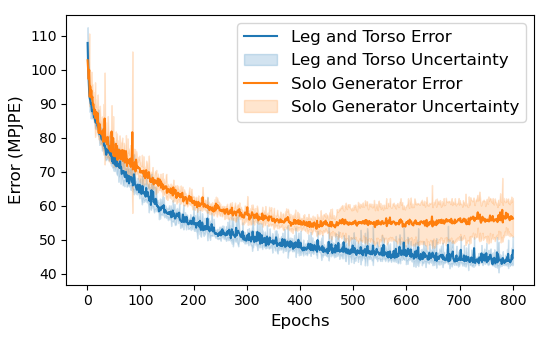}
\caption{The average evaluation error (MPJPE) and uncertainty of 5 leg and torso generators compared against 5 solo generators on the H36M validation dataset for each training epoch. Note how the leg and torso generators error is more stable during training where as the solo generators uncertainty increases once an optimum is found.}
\label{fig:erroroverepoch}
\end{figure}
One fundamental question when utilising unsupervised networks is when to stop training. However, we find a lack of information within prior work detailing how long authors have trained their models for. Therefore, we assume that prior work trained for a set amount of epochs and picked the weights across these epochs that performed best on an validation set. Though fine from an evaluation viewpoint, in practice this would not work if we had no ground truth data. In a truly unsupervised scenario there are three approaches we could use to decide when to stop training. Firstly, we could monitor the discriminators loss and stop training when it too weak or strong. Though there is intuition for this approach, in practice a strong discriminator can cause a generator to fail due to vanishing gradients \cite{Arjovsky2017TowardsPM} and a weak discriminator provides poor feedback to a generator reducing its performance. Secondly, we could visualise the predictions per epoch and decide by eye which pose is the best. Though having potentially hundreds of epochs and thousands of poses, this is not an efficient solution. Lastly, and most realistically, we could pick the final weight during the training of our model or average the weights between a certain range of epochs to use. For this scenario we show the stability of 5 solo and 5 independent leg and torso generators during adversarial training which can be seen in Figure \ref{fig:erroroverepoch}
As we can see, by having a leg and torso generator training independently not only is the MPJPE lower, but all our models converge to a similar minima during adversarial training. This is especially apparent around epoch 450 where our solo generators' errors begin to diverge away from one another. However, our leg and torso generators' error does not exhibit this same pattern and the errors are tighter to the mean during training following a stable downwards curve. As these models were trained for 800 epochs, if we chose the last epochs weights to evaluate on then the average error of the leg and torso generators' would have been $47.0\pm4.0$mm and our solo generators' average error would have been $56.6\pm5.8$mm. From epoch 500 to 800 the average error and standard deviation of our leg and torso generators' was $45.6\pm2.0$mm, the average error and standard deviation of our solo generators' by comparison was $56.3\pm5.5$mm. Taking the epoch where we observed the minimum error for each model would give us a result of $42.61\pm0.57$mm for the leg and torso generators' and $51.55\pm2.13$mm for the solo generators'.

\section{Conclusion}
To conclude, we present a rigorous study investigating how different 2D pose representations affects the unsupervised adversarial 2D-3D lifting process. Our results show that the most optimum pose representation for this process is that of an independent leg and torso which reduced the average error by 20\%. In fact, by having two lifting networks learn independently, we can improve both the accuracy and generalisability of a model. Furthermore, we have shown that when trained on a full pose, unintuitive correlations are induced between 2D coordinates and 3D ordinate predictions which we believe leads to poorer results. Additionally, we have shown that an independent leg and torso network can be recreated with greater consistency as they converge to similar minima during training whereas a full 2D pose approach diverges over time. However, the main statement from this paper is that nearly all of our different 2D pose representation networks improved upon one which used the standard full 2D pose representation. This highlights how simply optimizing the input to your model can significantly affect your results. Future work within this area could look at using attention mechanisms, which would allow the model to learn its own best representation for the 2D pose. Though for singular poses it may fall victim to the same problems as our full 2D pose representation model. In temporal scenarios, we believe it may surpass our independent leg and torso model as it would have the potential to represent the pose differently depending on what was required for that particular frame.
{\small
\bibliographystyle{ieee_fullname}
\bibliography{egbib}
}

\appendix

\end{document}